\documentclass{article}

\PassOptionsToPackage{numbers, compress}{natbib}

\usepackage[preprint]{neurips_2026}

\usepackage[utf8]{inputenc} 
\usepackage[T1]{fontenc}    
\usepackage{hyperref}       
\usepackage{url}            
\usepackage{booktabs}       
\usepackage{amsfonts}       
\usepackage{nicefrac}       
\usepackage{microtype}      
\usepackage{xcolor}         
\usepackage{graphicx}

\usepackage[nolist]{acronym}
\usepackage{bm}
\usepackage{amsmath}
\usepackage{multirow}
\usepackage{soul}
\usepackage{caption}
\usepackage{subcaption}
\usepackage[status=final,layout=inline]{fixme}

\title{Architecture-agnostic Lipschitz-constant Bayesian header and its application to resolve semantically proximal classification errors with vision transformers}

\author{
  Frederik Sch\"afer\\
  Institute of Structural Mechanics\\ and Dynamics in Aerospace Engineering,\\ University of Stuttgart\\
  Rockwell Collins Deutschland GmbH\\ 
  \texttt{frederik.schaefer@isd.uni-stuttgart.de}
  \And
  Luis Mandl\\
  Institute of Structural Mechanics\\ and Dynamics in Aerospace Engineering,\\ University of Stuttgart;\\
  Department of Hepatobiliary Surgery\\ and Visceral Transplantation,\\ University of Leipzig Medical Center\\
  \texttt{luis.mandl@isd.uni-stuttgart.de}
  \And
  Lars K\"alber\\
  Institute of Structural Mechanics\\ and Dynamics in Aerospace Engineering,\\ University of Stuttgart;\\
  \texttt{lars.kaelber@isd.uni-stuttgart.de}
  \And
  Tim Ricken\\
  Institute of Structural Mechanics\\ and Dynamics in Aerospace Engineering,\\ University of Stuttgart\\
  \texttt{tim.ricken@isd.uni-stuttgart.de}
}

\begin{document}

\begin{acronym}
    \acro{bnn}[BNN]{Bayesian neural network}
    \acro{nn}[NN]{neural network}
    \acro{vit}[ViT]{Vision Transformer}
    \acro{spce}[SPCE]{semantically proximal classification error}
    \acro{knn}[kNN]{k‑nearest neighbor}
    \acro{svd}[SVD]{singular value decomposition}
    \acro{kl}[KL]{Kullback-Leibler}
    \acro{resnet}[ResNet]{residual network}
    \acro{relu}[ReLU]{rectified linear unit}
    \acro{std}[STD]{standard-deviation}
    \acro{sn}[SN]{spectral norm}
    \acro{mc}[MC]{Monte-Carlo}
    \acro{lipbvit}[LipB-ViT]{Lipschitz-constant Bayesian header for vision transformer}
    \acro{bheadvit}[B-ViT]{Bayesian Header for Vision Transformer}
    \acro{pgd}[PGD]{projected gradient descent}
    \acro{fgsm}[FGSM]{fast gradient sign method}
    \acro{auc}[AUC]{are under the curve}
    \acro{pr}[PR]{precision-recall}
    \acro{roc}[ROC]{receiver operating characteristic}
    \acro{tsne}[t-SNE]{t-distributed stochastic neighbor embedding}
    \acro{elbo}[ELBO]{evidence lower bound}
\end{acronym}

\maketitle

\begin{abstract} 
  Label noise remains a critical bottleneck for the generalization of supervised deep learning models, particularly when errors are structured rather than random. Standard robust training methods often fail in the presence of such semantically proximal classification errors. This work presents an architecture-agnostic Lipschitz-constant Bayesian header that can be integrated into feature extractors such as vision transformers, yielding the bi-Lipschitz-constrained Bayesian Vision Transformer (LipB‑ViT). In contrast to conventional Bayesian layers, our approach enforces spectral normalization on both the mean and log-variance of the variational weights, which promotes calibrated predictive uncertainty and mitigates noise amplification. We further propose a novel metric to jointly capture uncertainty and confidence across misclassification rates, as well as an adaptive arithmetic-mean fusion scheme that combines feature-space proximity with predictive uncertainty to detect corrupted labels outperforming the state of the art k-nearest neighbor based identification methods by more than 7\%\ reaching a recall of more than 0.93 at 15\%\ semantically misclassified labels. Although computational costs increase due to Monte Carlo sampling, the method offers plug-and-play compatibility with pre-trained backbones and consistent hyperparameters across domains, suggesting strong utility for high-stakes applications with variable annotation reliability. The stabilized confidence estimates serve as the foundation for an analysis pipeline that jointly assesses dataset quality and label noise, yielding a second novel metric for their combined quantification. Lastly, we systematically evaluate LipB‑ViT under both structured (adversarial) and unstructured noise at inference time, demonstrating its robustness in realistic high-noise and attack scenarios. We compare its performance against state-of-the-art baseline methods.
\end{abstract}

\section{Introduction}
Although data quality is crucial for learning representations that generalize to real-world settings, the role of label noise has received comparatively little attention~\cite{Zha2025, Shi2024, Sambasivan2021}. In particular, label noise can induce optimization instabilities and significantly degrade generalization capabilities~\cite{zhang2017, northcutt2021neurips, Geiger2021}. Correction of noisy labels typically proceeds via three routes: (i) additional human effort to clean labels before training~\cite{Bernhardt2022, Northcutt2021, Pandey2022}, (ii) adaptations of the training pipeline to explicitly model annotator disagreement~\cite{Rodrigues2018, Guan2018}, or (iii) algorithmic robustness during training~\cite{Patrini2017, Muller2019, goldberger2017, han2018, Ren2018}. Human-in-the-loop label cleaning has been shown to incur substantial annotation costs. Moreover, it often remains inconclusive, since even experts disagree and may label space to accommodate edge cases, introducing downstream issues such as class imbalance. Annotator-aware architectures still require additional human effort in the form of multiple annotations, but can explicitly leverage disagreements in the labels. Robust-training approaches tolerate label noise without extra annotation cost by encoding ambiguity in the learning process, for example through loss correction~\cite{Patrini2017, Muller2019}, noise-adaptation layers~\cite{goldberger2017}, sample selection based e.g. on co-teaching~\cite{han2018}, or meta-learning~\cite{Ren2018}. However, these robust methods have a structural limitation: the guarantees typically only hold under symmetric and randomly distributed label noise. Nevertheless, empirical robustness is maintained even at noise rates as high as 80\%~\cite{Ghosh2017}. Once noise becomes structured, either class-dependent or concentrated near decision boundaries, the performance degrades significantly even at low noise rates, with existing mitigation methods providing little additional protection~\cite{Oyen2022}. In this work, we refer to this type of noise as \ac{spce}. One approach to modelling uncertainty in neural networks is the use of \acp{bnn}, which have been shown to improve the robustness and reliability of computer vision tasks in the context of uncertainty-aware segmentation and detection~\cite{kendall2017, gal2016, blundell2015}. While several Bayesian approaches to neural networks exist, including dropout~\cite{gal2016}, and (pseudo-)Bayesian deep ensembles~\cite{lakshminarayanan2017}, one prominent route is variational inference~\cite{blundell2015}. This approach places a distribution over each weight, rather than a single point estimate, and learns it by minimizing an objective that combines a data-likelihood term with a \ac{kl}-divergence regularizer~\cite{blundell2015}. Moreover, using only partial stochasticity in the network can retain useful uncertainty quantification while improving computational efficiency compared to fully Bayesian neural networks~\cite{sharma2023}, motivating architectures that employ Bayesian heads instead of fully Bayesian models. Furthermore, Lipschitz-constrained architectures that explicitly bound the rate of change in neural networks have been proposed to enhance stability and robustness by limiting output variation under small input perturbations~\cite{cisse2017}. Spectral normalization and related spectral bounding techniques have also been combined with \ac{bnn} to defend against adversarial test-time perturbations by softly regularizing an upper bound on the Lipschitz constant~\cite{Zhang2021}. This is achieved by adding an additional regularization term to the loss. We did not pursue this strategy further because we seek to avoid imposing Lipschitz constraints on the pre-trained feature extractor. Existing methods have so far mainly been demonstrated on comparatively small, fully Bayesian models trained from scratch, whereas the widespread use of \ac{vit}~\cite{dosovitskiy2021} and large pre-trained backbones would require a fully Bayesian treatment of all weights, which defeats the practical benefit of (re-)using pre-trained (deterministic) representations.

In this work, we address the problem of detecting noise on both sides of training: corrupted labels in the dataset that drive the well-known ``garbage-in, garbage-out'' failure mode, and corrupted or shifted inputs at inference time that erode a model’s reliability. The key contributions of this work are
\begin{enumerate}
    \item the introduction of an \textbf{architecture-agnostic Lipschitz-constant header}, that applies a hard spectral normalization to both the mean and log-variance of variational weights.
    \item a novel \textbf{misclassification severity metric} that quantifies uncertainty in the training dataset induced by label noise, enabling a universally applicable, cross-dataset metric to estimate the amount of label noise present.
    \item a method to \textbf{identify incorrect labels in the dataset} outperforming previous \ac{knn} based methods. 
    \item a thorough empirical \textbf{analysis of LipB‑Headers on Vision Transformers} on four open source datasets, comparing their robustness to \ac{spce} against state‑of‑the‑art baselines on label noise and post-training structured and unstructured noise. 
\end{enumerate}
Beyond these core contributions, we analyze the effect of \ac{spce} compared to random misclassification on (i) model accuracy and uncertainty and (ii) incorrect label identification.

\section{Methods}
The \ac{lipbvit} is added on top of a pre-trained \ac{vit} backbone and replaces its classification head with two stacked Bayesian linear layers, separated by a \ac{relu} activation function. The weights for these layers are sampled from learned Gaussian distributions parameterized by a mean and log-standard deviation, with a configurable prior~\cite{dosovitskiy2021}. We focus on \ac{vit}-16 for two reasons: widely used in both academic and industrial settings, and (ii) it has been reported to exhibit superior stability compared to traditional image classifiers~\cite{Gupta2023CertViT,dosovitskiy2021}. This work establishes a set of hyperparameters, which have been found to facilitate the use of the architecture as a plug-and-play replacement for a wide variety of models and data sets. In this work, the focus is exclusively on image classification, with the architecture under scrutiny. Given an input image $\mathbf{x}\in\mathbb{R}^{\rm{3\times H\times W}}$ and a pre-trained \ac{vit}-small 16 backbone $f_\theta:\mathbb{R}^{3\times H\times W}\!\to\!\mathbb{R}^{d}$, with height $\rm{H}$, width $\rm{W}$, and extracted feature space $\mathbf{z}$ with dimension $d$. The added Bayesian header $g_\phi$ then maps $\mathbf{z}$ to logits $\mathbf{o}\in\mathbb{R}^{C}$ through two stochastic linear layers with weight matrices $W^{(1)}, W^{(2)}\in\mathbb{R}^{m\times n}$ separated by a \ac{relu}, i.e., $\mathbf{o} = g_{\phi}(\mathbf{z}) = {W}^{(2)} \, \mathrm{ReLU}({W}^{(1)} \mathbf{z})$.

We chose a two-layer \ac{bnn} header because it exhibited the most stable behavior across hyperparameters and uncertainty estimates. This header is then parameterized by the variational parameters $\phi=\{\mu^{(\ell)},\rho^{(\ell)}\}_{\ell=1,2}$. In contrast to traditional neural networks layers there is no bias as affine offsets do not affect the \ac{sn}. The \ac{bnn} layer treats its weight matrix as a random variable with fully-factorized Gaussian variational posterior
\begin{equation}
q_\phi(W) = \prod_{i,j}\mathcal{N}\!\big(W_{ij}\,\big|\,\mu_{ij},\,\sigma_{ij}^{2}\big),
\qquad
\sigma_{ij}= \exp(\rho_{ij}),
\end{equation}
where $\rho_{ij}$ is unconstrained to enable $\sigma_{ij}>0$ automatically. We use the standard reparameterization trick to enable gradient-based optimization~\cite{kingma2015variational}. For the Gaussian prior, we place on every weight, ensuring an initial symmetric distribution around zero. We choose a small prior standard deviation of 0.01 for stability. At inference time, we draw 50 \ac{mc} samples and compute predictive uncertainty as the \ac{std} of the predicted class probabilities. We train the model by maximizing the standard \ac{elbo}, or equivalently, by minimizing the negative \ac{elbo}. For a dataset $\mathcal{D}=\{(\mathbf{x}_n,y_n)\}_{n=1}^{N}$ with mini-batch $\mathcal{B}\subset\mathcal{D}$ the loss function can then be defined as
\begin{equation}
\mathcal{L}(\phi)
=
\underbrace{\frac{1}{|\mathcal{B}|}\sum_{n\in\mathcal{B}}
 \mathbb{E}_{q_\phi}\!\big[\mathrm{CE}\!\big(g_\phi(\mathbf{z}_n),y_n\big)\big]}_{\text{data term}}
+
\beta \underbrace{\sum_{i,j}\!\Big[-\rho_{ij} + \log s_0+\frac{1}{2 s_0^{2}}\!\big(\exp{2\rho_{ij}}+\mu_{ij}^{2}\big) -\frac{1}{2}\Big]}_{\text{regularizer}},
\end{equation}
where $\beta = 10^{-4}$ controls the trade-off between fit and regularization with cross-entropy $\mathrm{CE}$, expectation under the variational distribution $\mathbb{E}_{q_\phi}$, and regularization factor $s_0$ as hyperparameter. The key novelty of the \ac{bnn} layer architecture is the use of spectral normalization over the entire weight distribution, which enforces a global upper bound on the Lipschitz constant of the header. This ``hard'' spectral bound mitigates outliers during \ac{mc} sampling by preventing high-norm weight realizations, while the spectral normalization in each $\tilde W^{(\ell)}$ enforces an upper bound on the Lipschitz constant of $g_\phi$. 

The deterministic version of the \ac{sn} is defined as
\begin{equation}
\|W\|_2 \;\equiv\; \sigma_{\max}(W)
= \max_{\mathbf{v}\neq 0}\frac{\|W\mathbf{v}\|_2}{\|\mathbf{v}\|_2}
 \sqrt{\lambda_{\max}(W^{SN}W)},
\end{equation}
with matrix $W\in\mathbb{R}^{m\times n}$ and where $\mathbf{v}\mapsto W\mathbf{v}$ is a linear map on $\|W\|_{2,\rm{Lipschitz}}$. Since computing $\sigma_{\max}$ via \ac{svd} is too expensive at every step, the layer keeps a persistent left singular vector buffer $\mathbf{u}\in\mathbb{R}^m$ and refines it with $SN$ power iterations per forward call. While multiple iterations improve convergence, empirical results suggest that one iteration per step, often suffices due to batch-wise refinement along the same Krylov subspace. Unlike standard spectral normalization, which constrains a single weight matrix, our approach keeps the entire variational distribution spectrally bounded. The implementation does this by re-scaling both $\mu$ and $\sigma$ by the same factor $\hat\sigma$
\begin{equation}
\widetilde{W} \tilde{\mu} + \tilde{\sigma}\odot\bm{\varepsilon}
 \frac{1}{\hat{\sigma}+\varepsilon}\big(\mu + e^{\rho}\odot\bm{\varepsilon}\big)
 \frac{W}{\hat{\sigma}+\varepsilon},
\end{equation}
where $\tilde{(\cdot)}$ denotes quantities prior to normalization, and all remaining quantities are estimated via power iterations. This then leads to a shape invariance as the same scalar divides $\mu$ and $\sigma$, where the entire variational distribution is just isotropically shrunk and relative uncertainty is preserved. Here $\lVert \tilde{W} \rVert_2 \approx 1$ for stable $\hat{\sigma}$ ensures an approximate 1-Lipschitz constraint on the layer. With both layers spectral normalized, the header is approximately 1-Lipschitz. The header here only strictly enforces the Lipschitz upper bound on the head, not full bi-Lipschitzness end-to-end~\cite{liu2020sngp}. Bi-Lipschitzness ensures a distance-preserving mapping up to constant factors, ruling out both unbounded expansion and degenerative contraction of distances. The lower Lipschitz bound is implicitly provided by the \ac{vit} backbone, assumed to be non‑collapsing due to pre-training, together with a hidden layer in the header. In practice, the estimator becomes distance-aware as the spectral norm caps the growth of logit variance, while the variational posterior in combination with a weak prior preserves epistemic uncertainty in unconstrained directions. Consequently, the \ac{mc} variance tracks the distance from the training data manifold rather than being arbitrarily amplified by deep-net nonlinearities. Spectral normalization is applied to the entire weight distribution, thereby enforcing a global Lipschitz bound. In contrast, post-hoc normalization of individual MC samples leaves the distribution’s support unbounded and susceptible to sporadic high-norm realizations induced by uncontrolled weight-norm fluctuations. The proposed scheme preserves the Gaussian form required by the variational \ac{kl}-divergence, maintains a scale-invariant signal-to-noise ratio, and leads to stable gradients as well as more efficient inference.
\subsection{Training methods}
We compare two training methods for handling label noise: (i) standard training with a single network, and (ii) a co-teaching approach~\cite{han2018}. Co-teaching utilizes two neural networks in parallel, where each selects instances with a small loss, which are considered clean, from its mini-batch. These instances are then exchanged with the peer network for parameter updates. The objective of this method is to mitigate confirmation bias by decoupling the selection of clean samples from model updates, using a dynamic, dropout-like mechanism that retains only the most reliable training samples. To achieve this, an important hyperparameter is the assumed noise rate $\tau \approx \epsilon$. For a deterministic, epoch-scheduled forget rate $R(t) = 1 - \tau \cdot \min(t/t_k, 1)$ that starts at 0 and linearly ramps up over the first $t_k$ epochs to the assumed noise rate $\tau \approx \epsilon$. However, in the absence of prior knowledge about the dataset’s label-noise rate, this strategy proved to be suboptimal, motivating the incorporation of an adaptive forget rate. The forget rate is determined on a per-mini-batch basis as a function of the number of samples for which the two networks yield discordant predictions. Per mini-batch $\mathcal{B}$, the forget rate is set adaptively using the predictions $\hat{y}_f$ and $\hat{y}_g$ of the peer networks as
\begin{equation}
R_{\text{forget}}(\mathcal{B}) = \min\!\left(R_0 + \alpha \cdot \frac{1}{|\mathcal{B}|}\sum_{x \in \mathcal{B}} \!\left[\hat{y}_f(x) \neq \hat{y}_g(x)\right], R_{\max}\right),
\end{equation}
with $R_0 = 0.1$, $\alpha = 0.2$, and $R_{\max} = 0.5$.
\subsection{Metric to evaluate model uncertainty determinability and stability over misclassification}
A tailored metric is needed to evaluate the reliability and stability of uncertainty and confidence estimates under miscalibration. To accurately assess uncertainty and confidence in the misclassification rate, a simple averaging over random seeds and datasets is insufficient, as it ignores variability across runs. Both changes in the mean and the associated standard deviation or broader dispersion measures must be taken into account. Consequently, we designed a metric that explicitly accounts for both mean trends and variability across runs. The emphasis is on obtaining a robust estimate of dataset quality from uncertainty and confidence values. The extent to which an estimate of the quality of the dataset is both reliable and precise is a pivotal consideration. In order to achieve this objective, a Gaussian response model is fitted to the repeated measurements over the seeds at each misclassification rate, yielding a posterior distribution for each observation. This results in two performance metrics: (i) the soft confidence defined as the posterior probability mass within a range of $\pm 2\%$ of the actual misclassification averaged over all misclassification rates; and (ii) the soft accuracy, defined as the proportion of observations for which the maximum a posteriori estimate fell within the same neighborhood of $\pm 2\%$. Together, these metrics quantify both probabilistic certainty and discrete prediction reliability under a tolerance for near-miss assignments.

\subsection{Feature extractor and suspicion score}
For the \ac{knn}-based suspicion score, we employed a \ac{resnet}-50 model pre-trained on ImageNet, with its classification head removed, as a feature extractor~\cite{he2016deep}. From this network, we extracted 2048-dimensional feature vectors and computed suspicion scores directly within this feature space using \acp{knn}. The model was used in its pre-trained state without fine-tuning ~\cite{liu2020sngp, papernot2018dknn}. For each sample $i$ with \ac{resnet}-50 feature $\mathbf{z}_i\in\mathbb{R}^{2048}$ and assigned label $y_i$, the $K{=}10$ cosine-nearest neighbors $\mathcal{N}_K(i)$ with distances $d_{i,k}$ and labels $y_{j_{i,k}}$ are calculated. Using the distance agreement ($w_{i,k}\propto 1/(d_{i,k}+\epsilon)$, $\sum_k w_{i,k}=1$), we obtain $a_i = \sum_{k=1}^{K} w_{i,k}\,\{y_{j_{i,k}}=y_i\}\;\in\;[0,1]$. These values are then normalized to $[0,1]$ and the suspicion score is calculated as $s^{\text{kNN}}_i = 1 - \hat a_i\;$.

\subsection{Uncertainty based suspicion score}
Predictive uncertainty is quantified via \ac{mc} sampling~(50 forward passes through the \ac{lipbvit}). For each sample, the uncertainty score is defined as the variance $\sigma_{s}$ of the predicted-class $c$ probability $p$ with $u_i = \sqrt{\sigma_{s}^2\,p^{(s)}_{\hat c_i}}$ with number of \ac{mc} samples $s$, and subsequently min-max normalized to $[0,1]$. The resulting quantity can be interpreted as an uncertainty-based suspicion score.

\subsection{Adaptive arithmetic-mean fusion}
A naive fusion strategy would be to take the unweighted arithmetic mean of both scores, which leaves threshold-independent metrics unchanged. However, to facilitate threshold selection, this is converted into an unsupervised method with a user-supplied class prior in the form of an expected misclassification rate within the dataset. To achieve this, each rank distribution is scored using Sarle's bimodality coefficient~\cite{sarle1985modeclus} using a slope of $\beta=8$. With the expected misclassification rate, acting only a calibration factor. We furhter clip the weights at ($\in[0.2,0.8]$) ensure a balanced fusion by preventing either signal from dominating completely or being ignored.
\subsection{Semantically proximal classification errors}
The notion of semantic misclassification is based on the assumption that visually similar pairs of samples have a substantially higher probability of being confused by the classifier. Conversely, samples that are clearly distinct are expected to show markedly lower misclassification rates. We use \ac{resnet}‑50 as a feature extractor, by removing the final classification layer and taking the 2048‑dimensional output after the global average pooling layer to extract the image features. Utilizing the extracted features, we apply \ac{knn} to identify samples that are closely related in the feature, but belong to different classes. Then their labels are swapped, achieving synthetically generated \acp{spce} by changing the labels of samples that lie closer to other class clusters so that they are relabeled according to these neighboring clusters.
\section{Datasets and experimental setup}
To evaluate the performance of \ac{lipbvit} and the baseline methods, we employ four publicly available benchmark datasets, summarized in Table~\ref{tab:datasets_clean}. These datasets comprise two medical imaging tasks, based on magnetic resonance imaging (MRI) and histopathology, and two industrial inspection tasks involving magnetic tiles and steel surfaces. In all four scenarios, human experts routinely perform the visual assessment, yet misclassifications can occur and may entail substantial clinical or economic consequences, underscoring the need for reliable and precise automated decision support.
\begin{table}[h]
 \caption{Datasets used for comparison with number of classes $C$, number of images used in training $N_{\text{train}}$, number of images used in testing $N_{\text{test}}$, and image size in pixels.}
  \label{tab:datasets_clean}
  \centering
  \begin{tabular}{lccccl}
    \toprule
    \textbf{Dataset} & \textbf{$C$} & \textbf{$N_{\text{train}}$} & \textbf{$N_{\text{test}}$} & \textbf{Size} & \textbf{Source}\\
    \midrule
    Brain Tumor MRI &
    4 &
    $5{,}712$ &
    $1{,}311$ &
    $224^2$ &
    \citet{msoud_nickparvar_2026}\\

    Colorectal Histology &
    8 &
    $4{,}001$ &
    $1{,}002$ &
    $224^2$ &
    \citet{Kather2016}\\

    Magnetic Tile Defects &
    6 &
    $1{,}069$ &
    $276$ &
    $224^2$ &
    \citet{huang2020surface}\\

    Steel Surface Defects &
    6 &
    $1{,}440$ &
    $360$ &
    $224^2$ &
    \citet{cao2025neu_cls}\\
    \bottomrule
  \end{tabular}
\end{table}
We evaluate the models under both pre-training and post-training structured and unstructured noise. To study the impact of structured and unstructured label noise during pre-training, classifiers are evaluated on four image-classification benchmarks under two label-corruption regimes (random and semantic) with corruption rates $\eta \in [1\%,15\%]$. Each setting (dataset, regime, $\eta$, model) is instantiated with five random seeds, and performance is always measured on a clean validation set to ensure fair comparison. In the post-training evaluation phase, individual noise types are added to the images with gradually increasing intensity. The considered corruptions include Gaussian noise, shot noise, read noise, thermal and sensor noise, Gaussian and motion blur, brightness variation, and demosaicing artifacts. Structured noise is modeled via adversarial attacks. Specifically, we repeat the steps described above, but instead of applying different noise sources, we generate adversarial perturbations with gradually increasing strength. For a fair and effective evaluation, we use a ResNet‑50 model fine‑tuned on the respective dataset as the base (attacker) model when crafting adversarial examples, since attacks are generally more effective when the target model closely matches the attacker model. A categorization of attacks can be made into two types: \ac{fgsm}~\cite{goodfellow2015explaining} and \ac{pgd}~\cite{madry2018towards}. The experiments benchmark \ac{lipbvit} against (i) a standard \ac{vit} (deterministic baseline) and (ii) a \ac{bheadvit} with identical architecture but without spectral normalization. A fully \ac{bnn} was excluded due to inferior performance on pre-training-dependent datasets.
\section{Results}
The uncertainty framework is evaluated against state-of-the-art methods across four primary dimensions: (i) robustness to label noise during training, (ii) capability to provide an estimate on the data quality, (iii) capability to identify incorrect labels (misclassification detection), and (iv) robustness to input perturbations and adversarial attacks at inference.
\subsection{Robustness to label noise}
We evaluate three model types \ac{vit}, \ac{bheadvit} and \ac{lipbvit} with two different training approaches. Furthermore we evaluate the \ac{lipbvit} for $SN=1$ and $SN=5$ number of power iterations. 
\begin{figure}[h]
    \centering
    \includegraphics[width=5.5in, keepaspectratio]{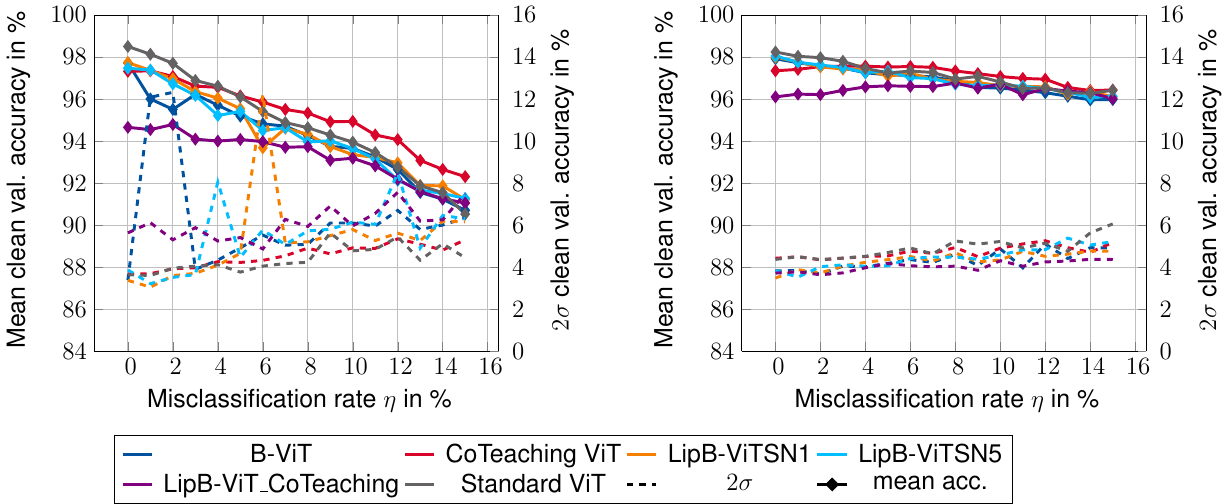}
    \caption{Training accuracy over misclassification rate for all compared models. The solid lines represent the accuracies, while the dotted lines indicate the $\pm2\sigma$ standard deviation. On the left are the results for \ac{spce}, and on the right are those for random label noise.}
    \label{fig:fig_mean_clean_misclassification}
\end{figure}
As shown in Figure~\ref{fig:fig_mean_clean_misclassification}, the accuracy degradation for a random misclassification is on average only 1.46\% across all models, with a negligible performance difference of less than one percent between models. In contrast, the average accuracy drop over all models for the \ac{spce} is 6.2\%. Table \ref{tab:model_noise_robustness} summarizes the relative accuracy drop and the stability across all four datasets and the five random seeds per dataset.
\begin{table}[h]
  \caption{Model robustness comparison under increasing \ac{spce}. Reported are the initial performance at $\eta=0\%$, final performance at $\eta=15\%$, and the relative drop in accuracy.}
  \label{tab:model_noise_robustness}
  \centering
  \begin{tabular}{lccc}
    \toprule
    \textbf{Model} & \textbf{$\eta=0\%$} & \textbf{$\eta=15\%$} & \textbf{Drop (\%)} \\
    \midrule
    \ac{bheadvit} & $97.7 \pm 1.7$ & $90.7 \pm 3.2$ & 7.2 \\
    CoTeaching \ac{vit} & $97.3 \pm 1.8$ & $\mathbf{92.3 \pm 2.6}$ & 5.2 \\
    \ac{lipbvit}SN1 & $97.7 \pm 1.7$ & $91.2 \pm 3.1$ & 6.7 \\
    \ac{lipbvit}SN5 & $97.5 \pm 2$ & $91.3 \pm 3.2$ & 6.3 \\
    \ac{lipbvit}\_CoTeaching & $94.7 \pm 2.9$ & $91.1 \pm 3.8$ & \textbf{3.8} \\
    Standard \ac{vit} & $\mathbf{98.5 \pm 1.8}$ & $90.5 \pm 2.2$ & 8.1 \\
    \bottomrule
  \end{tabular}
\end{table}
The standard \ac{vit} model demonstrates the highest initial performance, with an accuracy of $98.5\% \pm 1.8\%$. However, it also exhibits the lowest accuracy, with a misclassification rate at $\eta=15\%$ of $90.5\% \pm 2.2\%$ that is a drop of about 8\%. The highest accuracy at $\eta=15\%$ misclassification was achieved by the \ac{vit} with our adaptive coaching approach at $92.3\%\pm 2.6\%$ and an average accuracy drop of 5\%. Concerning the mean accuracy decline when employing conventional training methods, the \ac{lipbvit} demonstrates marginal inferiority, exhibiting an accuracy decline of 6.3\% across all datasets. The lowest accuracy drop but also lowest initial performance is shown by the \ac{lipbvit}\_CoTeaching model. 
\subsection{Data quality metric}
Based on confidence- and uncertainty-weighted misclassification rates, we propose a novel data-quality metric that estimates the number of potential misclassifications in a dataset. We compare this metric against radius-based uncertainty measures in the \ac{vit} feature space. To assess how accurately and robustly model uncertainty reflects true errors, we introduce a dedicated evaluation metric that quantifies both the correctness of uncertainty estimates and their stability with respect to misclassifications. This metric $P(\lvert \hat{x} - x_{\text{true}} \rvert \le 2) \approx \text{soft\_accuracy}\_{w=2}$ captures the accuracy and confidence of predicting the data quality metric for a new sample. The corresponding results are reported in Table~\ref{tab:soft_metrics}. 
\begin{table}[h]
  \caption{Soft accuracy and confidence metrics under uncertainty- and confidence-based weighting.}
  \label{tab:soft_metrics}
  \centering
  \begin{tabular}{lcccc}
    \toprule
    \textbf{Model} 
      & \multicolumn{2}{c}{\textbf{Soft Acc.}} 
      & \multicolumn{2}{c}{\textbf{Soft Conf.}} \\
    \cmidrule(lr){2-3} \cmidrule(lr){4-5}
      & \textbf{(Unc.)} & \textbf{(Conf.)} 
      & \textbf{(Unc.)} & \textbf{(Conf.)} \\
    \midrule
    \ac{bheadvit}            & 0.47 & 0.46          & 0.41 & 0.37          \\
    \ac{lipbvit}SN1          & 0.43 & \textbf{0.53} & 0.40 & \textbf{0.42} \\
    \ac{lipbvit}\_CoTeaching & 0.32 & 0.40          & 0.34 & 0.34          \\
    \ac{lipbvit}SN5          & 0.40 & 0.49          & 0.39 & 0.41          \\
    Radius Based             & 0.26 & --            & 0.34 & --            \\
    Standard \ac{vit}                 & --   & 0.42          & --   & 0.38          \\
    \bottomrule
  \end{tabular}
\end{table}
We compare the fitted data quality metric with respect to its ability to determine unambiguity and stability under misclassification. Specifically, we evaluate both $1 - \text{prediction confidence}$ and the uncertainty. Among the approaches considered, the radius-based uncertainty yielded the lowest performance, followed by the Standard \ac{vit}. The best results are obtained with the \ac{lipbvit}SN1, which interestingly shows higher performance in terms of confidence over misclassification than uncertainty over misclassification. This finding indicates that the Bayesian spectrally bounded head not only regularizes uncertainty estimates but also stabilizes prediction confidence, which is defined as the predictive mean over 50 \ac{mc} samples.
\subsection{Capability to identify incorrect labels}
\begin{figure}[h]
    \centering
    \begin{subfigure}[b]{0.49\textwidth}
        \centering
        \includegraphics[width=\linewidth, keepaspectratio]{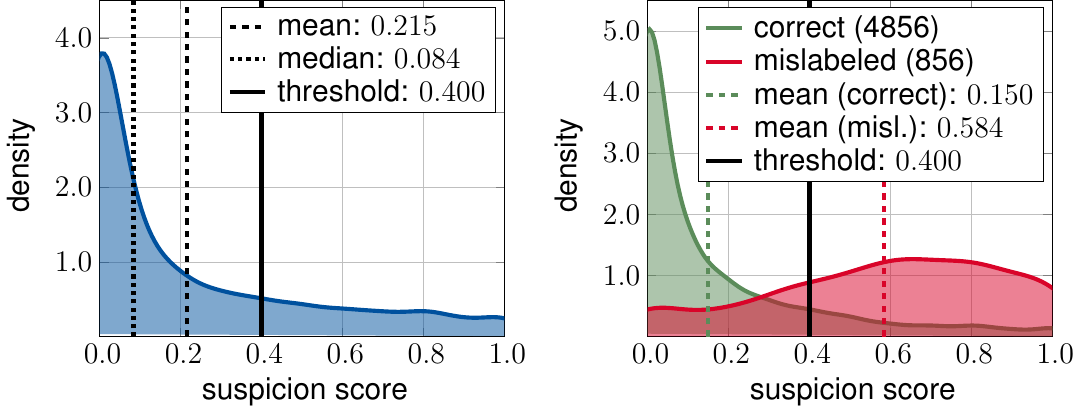}
        \caption{\ac{knn}-based method.}
        \label{fig:fig_knn_only_analysis}
    \end{subfigure}
    \hfill
    \begin{subfigure}[b]{0.49\textwidth}
        \centering
        \includegraphics[width=\linewidth, keepaspectratio]{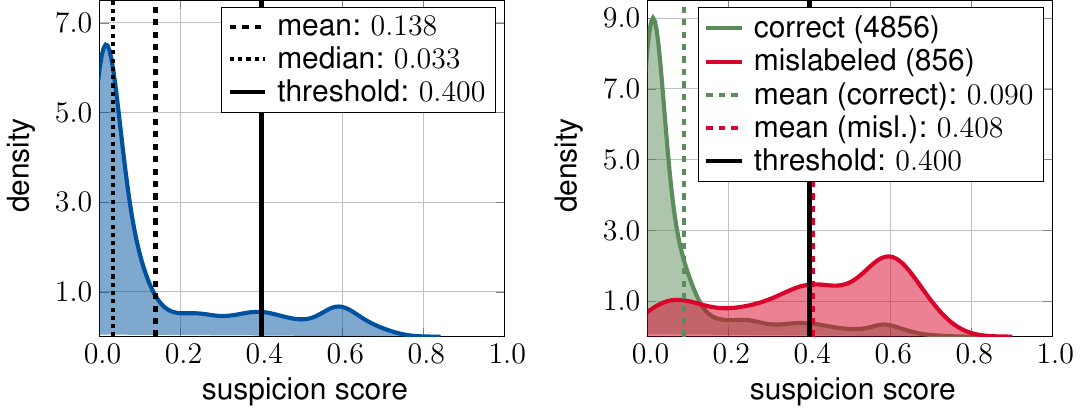}
        \caption{Adaptive arithmetic-mean fusion method.}
        \label{fig:fig_bayes_posterior_analysis}
    \end{subfigure}
    \caption{Suspicion-score densities for the two methods. Each panel shows on the left the blind marginal density $p(s)$ over all samples, and on the right the ground-truth-conditional densities $p(s\mid\text{correct})$ and $p(s\mid\text{mislabeled})$, whose separation indicates how well the score discriminates label noise.}
    \label{fig:density_comparison}
\end{figure}

The comparison of suspicion scores in Table \ref{tab:misclassification_performance} illustrates the necessity of fusing multiple signals. The adaptive arithmetic-mean fusion method consistently outperforms the standalone \ac{knn} and uncertainty method. For the adaptive arithmetic-mean fusion-based suspicion score, we did not pre-train the feature extractor; for further details, see the appendix.
\begin{table}[h]
  \caption{Performance comparison under different misclassification types using \ac{auc}-\ac{pr} and \ac{auc}-\ac{roc} metrics (mean $\pm$ standard deviation).}
  \label{tab:misclassification_performance}
  \centering
  \begin{tabular}{llcc}
    \toprule
    \textbf{Misclassification Type} & \textbf{Method} & \textbf{\ac{auc}-\ac{pr} Mean $\pm$ Std} & \textbf{\ac{auc}-\ac{roc} Mean $\pm$ Std} \\
    \midrule
    Random & \ac{knn} & $0.83 \pm 0.19$ & $0.99 \pm 0.04$ \\
    Random & uncertainty & $0.26 \pm 0.08$ & $0.81 \pm 0.06$ \\
    Random & Bayes & $\mathbf{0.88 \pm 0.14}$ & $\mathbf{0.99 \pm 0.03}$ \\
    \ac{spce} & \ac{knn} & $0.51 \pm 0.25$ & $0.92 \pm 0.05$ \\
    \ac{spce} & uncertainty & $0.28 \pm 0.10$ & $0.84 \pm 0.06$ \\
    \ac{spce} & Bayes & $\mathbf{0.55 \pm 0.20}$ & $\mathbf{0.94 \pm 0.04}$ \\
    \bottomrule
  \end{tabular}
\end{table}
The Adaptive arithmetic-mean fusion method demonstrates improved separability in density scores, facilitating a more defined data cutoff line as shown Figure~\ref{fig:fig_bayes_posterior_analysis} compared to Figure~\ref{fig:fig_knn_only_analysis}.
Adaptive arithmetic-mean fusion combines the structural information of the feature space with the sampled based uncertainty of the \ac{lipbvit}. This fusion yields a higher \ac{auc}-\ac{pr}, indicating that the method more effectively separates correctly and incorrectly labeled instances.
\subsection{Robustness to input perturbations and adversarial attacks at inference}
\begin{figure}[h]
    \centering
    \includegraphics[width=5.5in, keepaspectratio]{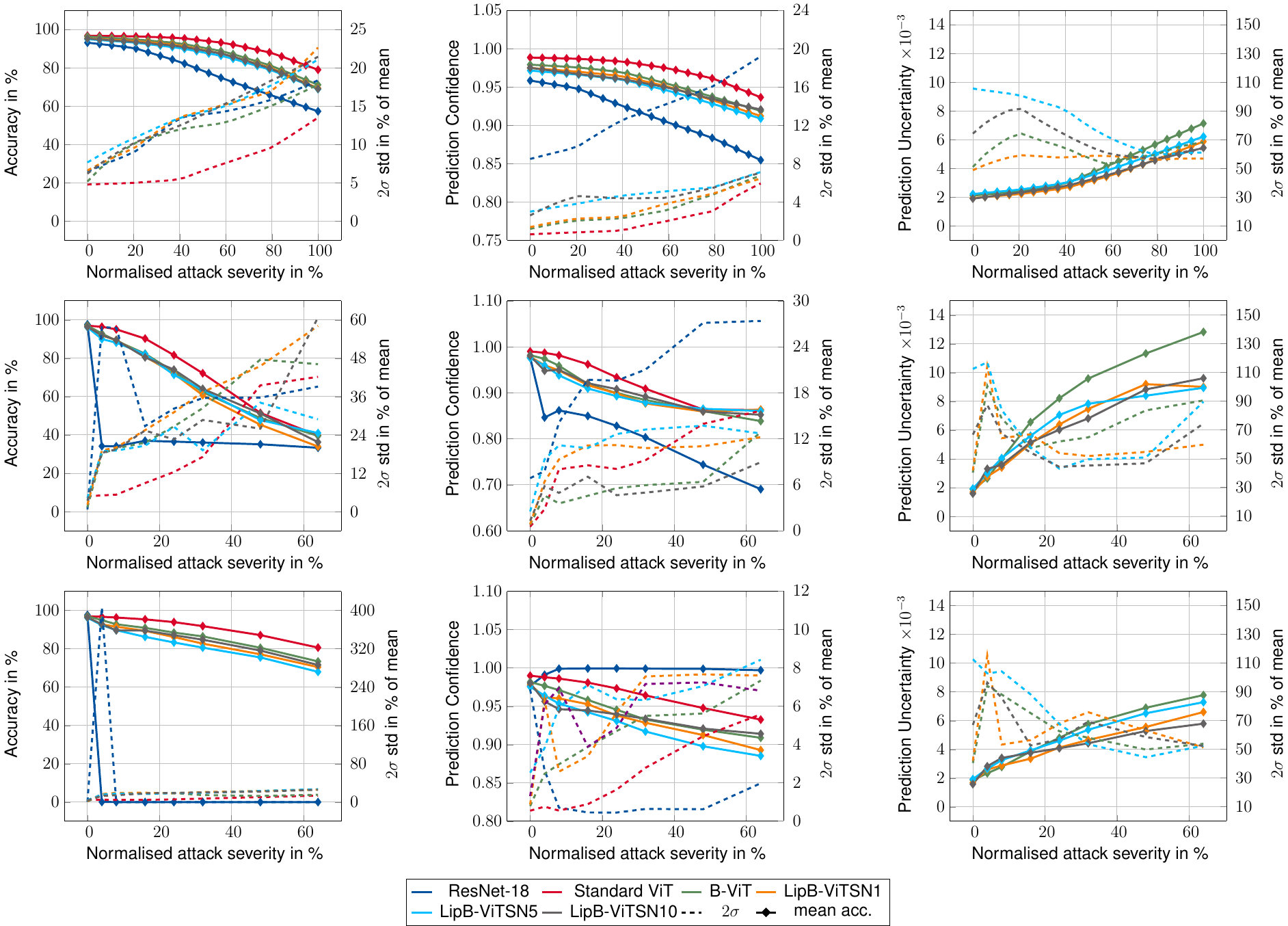}
    \caption{Accuracy, prediction confidence, and predictive uncertainty ($\times{10^{-3}}$) under unstructured image corruptions~(top), \ac{fgsm}~(middle), and \ac{pgd}~(bottom) transfer attacks sourced from \ac{resnet}-18, pooled across four datasets and five seeds. Solid lines denote mean performance; dashed lines indicate $2\sigma$ inter-seed \ac{std} as a percentage of the mean (right y-axis).}
    \label{fig:attack}
\end{figure}
Figure~\ref{fig:attack} summarizes the averaged results: unstructured corruptions are shown in the top row, while \ac{fgsm} and \ac{pgd} attacks appear in the middle and bottom rows, respectively. The \acp{vit} accuracy is the least affected, with a decrease of only approximately 18\%. The strongest decrease is observed in the \ac{resnet}. The discrepancies in accuracy between the Bayesian models for the three noise types are found to be negligible. For unstructured noise, the accuracy drops by about 27\%.
The results for all noise types for the \ac{bheadvit} compared to the \ac{lipbvit} are negligible in terms of accuracy and confidence. The main difference is that the Lipschitz regularization substantially reduces inter-seed and inter-dataset uncertainty \ac{std} under corruption. 

\section{Discussion}
Despite improvements, some constraints of the Bayesian header remain. The main issue is the significant computational overhead compared to standard deterministic models, which arises from the need for \ac{mc} sampling and multiple power iterations. While synthetic label swaps effectively model class confusion, real-world noise includes additional complexities, such as inconsistent annotation criteria or multi-modal errors, that our approach does not replicate. This limits the generalizability of our findings to practical settings. Moreover, the experiments were conducted on relatively small datasets with a fixed input resolution of $224^2$; the scalability of the Lipschitz constraint to very large-scale, ImageNet-sized datasets remains to be verified, as the memory footprint of storing singular vectors for spectral normalization could become a bottleneck. The misclassification severity metric shows dataset-dependent horizontal shifts, likely due to variations in noise distributions or baseline label accuracy. Robust validation would require evaluation on larger, expert-annotated datasets to disentangle these effects. This concern is supported by our Bayesian posterior fusion analysis, which identified a mislabeled sample in the brain tumor dataset that was subsequently confirmed by an expert: the same image appeared in both its original and mirrored form, but the two versions were assigned to different classes (see appendix). Consequently, residual label errors in the baselines used for evaluation may have introduced a small shift in our metric and an artificial widening of the standard deviation, making the misclassification severity appear less sharp than it actually is. Additionally, our \ac{knn}-based suspicion score relies on a non-tuned feature extractor. Although the literature predominantly advises against fine-tuning under label noise, some studies suggest that fine-tuning a model for a few epochs can improve the feature structure before learning the noise, which could, in principle, enhance the \ac{knn}-based suspicion score. However, for a new dataset, it is unclear how to choose an appropriate limit for ``a few'' epochs. A more detailed analysis of feature extractors fine-tuned on noisy datasets is provided in the appendix; in our experiments, fine-tuning the feature extractor decreased the performance of the \ac{knn}-based suspicion score by an average of about 14\%. The ``plug-and-play'' nature of the proposed header is a significant advantage. As shown in our results, the architecture is agnostic to the backbone (here tested on \ac{vit}‑16, but in principle applicable to other architectures). The fact that the same hyperparameters yielded similar results across medical and industrial datasets suggests that the method is robust across domains, making \ac{lipbvit} a strong candidate for scenarios in which data collection is expensive and expert annotation is prone to error.
\section{Conclusion}
In this work, we addressed the dual challenges of label noise during training and input corruption during inference in \ac{vit}-based models. While random label noise is known to degrade performance, our experiments demonstrate that \ac{spce} induce a substantially sharper decline in accuracy, dropping model performance by over 6\% on average compared to roughly 1.5\% for random corruption. By integrating a Lipschitz-constrained Bayesian head, \ac{lipbvit} not only mitigated the accuracy drop under \ac{spce} but also produced more consistent uncertainty estimates across all datasets. Leveraging this knowledge, we introduced a novel metric derived from the predictive uncertainty to quantify data quality. This approach allows for a cross-dataset estimate of label noise levels. We further introduced a pipeline based on Bayesian fusion of feature-based \ac{knn} scores and model-derived uncertainty to isolate mislabeled instances. This pipeline demonstrated a marked improvement in performance when compared with the \ac{knn}-based label-noise detection method. It reliably identified and isolated incorrect labels in the dataset for both the random and \ac{spce} cases. Lipschitz regularization enhanced robustness to post-hoc perturbations, preserving the utility of uncertainty estimates under both random and adversarial input corruptions during inference.
\fxwarning{STD vs sgima vs variacne}
\bibliography{neurips}

\newpage

\clearpage
\appendix
\section{Software and Hardware}
Experiments were conducted with Python 3.11.11 (CPython) on a Linux system (6.8.0-106-generic) using PyTorch 2.5.1 (CUDA 12.4, cuDNN 90100) and key libraries: NumPy 2.2.1, SciPy 1.15.1, scikit-learn 1.6.1, and TIMM 1.0.14. The hardware setup included a 48-core/96-thread AMD EPYC 9474F CPU with 755 GB RAM and two NVIDIA RTX 6000 Ada GPUs (47.4 GB VRAM each, compute capability 8.9). The average per epoch training time depending on the dataset ranges from 14-32 seconds. The optimizer used was AdamW from PyTorch.

\section{Supplementary plots}
\begin{figure}[t]
    \centering
    \includegraphics[width=5.5in, keepaspectratio]{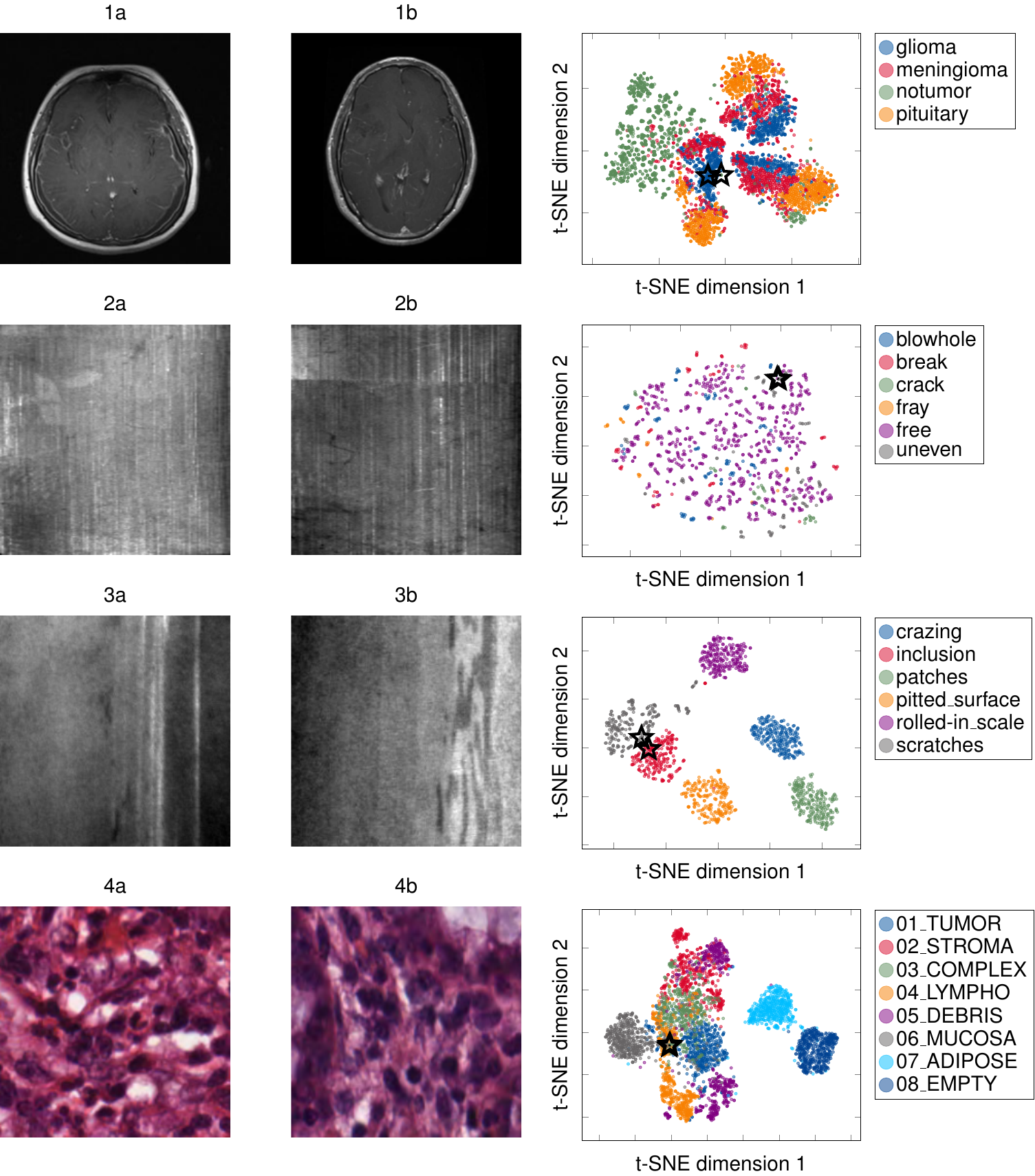}
    \caption{This plot shows semantically close samples from each dataset with different classes, introducing what it is easier to miscategorize samples that are similar than those with obvious differences. On the left side, the images are always shown first, and on the right side, a \ac{tsne} plot of the \ac{resnet} reduced features is displayed, with the samples shown in the images marked. Row 1: Dataset - Brain tumor 1a: No tumor, 1b: Glioma. Row 2: Dataset - Magnetic tile surface defects 2a: Free, 2b: Uneven. Row 3: Dataset - Steel surface defects 3a: Scratches, 3b: Inclusions. Row 4: Dataset - Colorectal 4a: Complex, 4b: Mucosa. This visualization highlights how semantically similar samples can lead to easier mislabeling.}
    \label{fig:whynofintuningfeatureextractor}
\end{figure}

\begin{figure}[t]
    \centering
    \includegraphics[width=5.5in, keepaspectratio]{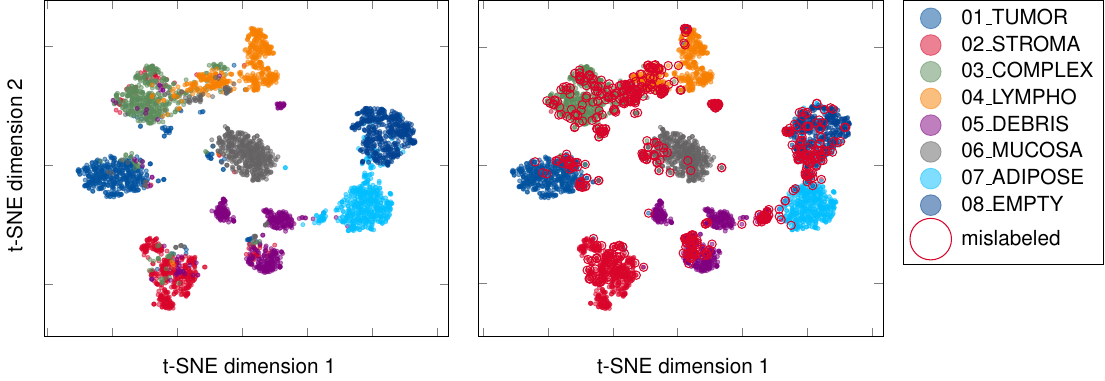}
    \caption{This plot displays the reduced features from \ac{resnet}, further reduced via \ac{tsne}. The \ac{resnet} was trained for 40 epochs, achieving the highest accuracy on a clean validation set. The training data contained semantically miscslassefied samples. On the left, the \ac{tsne} features are shown with their original (not miscslassefied) classes, clearly showing label noise merging into the class clusters. On the right, the miscslassefied samples are marked, showing a near-perfect alignment between cluster noise and miscslassefied samples. This demonstrates that even when validation accuracy on a clean set is optimal, the feature extractor learns the miscslassefied patterns.}
    \label{fig:fig_no_fine_tuning}
\end{figure}

\begin{figure}[t]
    \centering
    \includegraphics[width=5.5in, keepaspectratio]{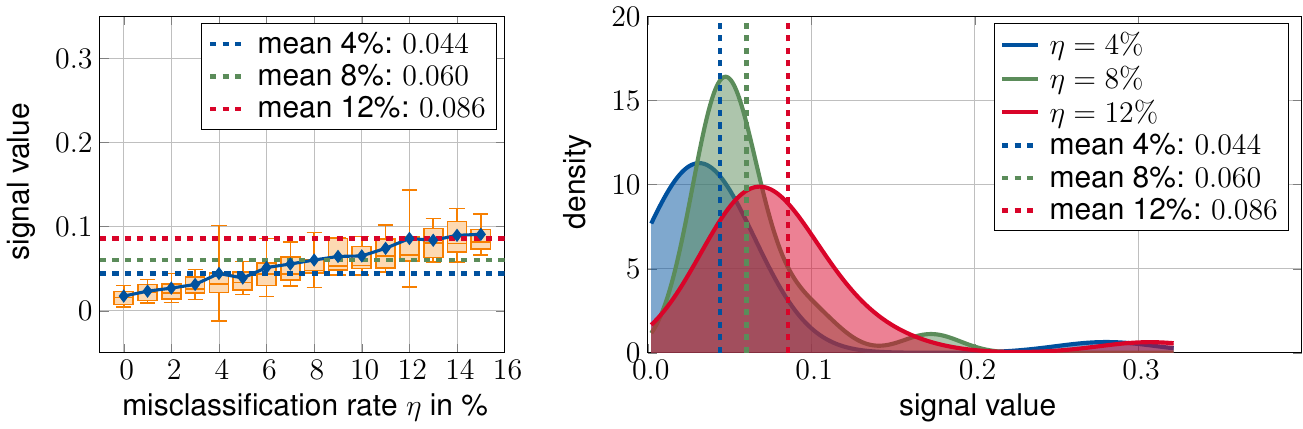}
    \caption{This figure shows on the left the inverse confidence of the \ac{lipbvit} over the misclassification rate. The plot on the left combines, as an interquartile range bar plot, the information from all 5 seeds and over all 4 datasets, considering both random and \ac{spce} methods. Furthermore, the dotted lines represent cross-sections at 4\%, 8\%, and 12\%, where each line intersects the mean, demonstrating the possibility of identifying a new value and tracing it back to a misclassification in the dataset. On the right, the distributions at given means are shown.}
    \label{fig:fig_inv_confidence_lipbvit}
\end{figure}
\begin{figure}[t]
    \centering
    \includegraphics[width=5.5in, keepaspectratio]{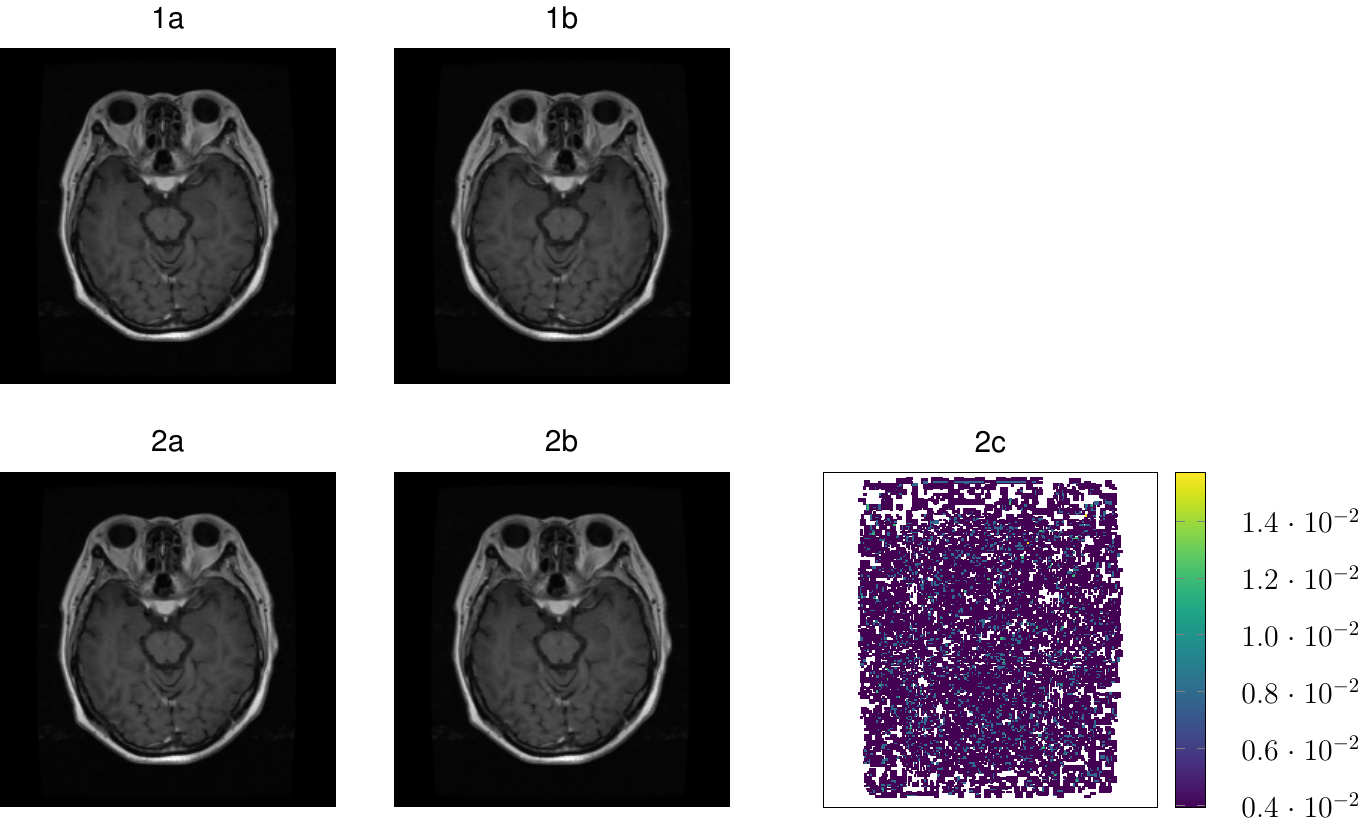}
    \caption{This plot shows an example where our pipeline actually marked an expert-verified false label in the original dataset. On the top row, the original images are shown: 1a is defined as class meningioma, and 1b is pituitary. On the second row, the first image (2a) is the same as 1a (1a = 2a). However, 2b is the mirrored version of 1b (1b mirrored = 2b). The mean absolute error between the pixel values in the images is shown, and the difference is nearly within machine precision. Therefore, the same image, when mirrored, was assigned to two classes. This demonstrates the effectiveness and necessity of our pipeline.}
    \label{fig:fig_brain_tumor_thresh}
\end{figure}
\begin{figure}[t]
    \centering
    \includegraphics[width=5.5in, keepaspectratio]{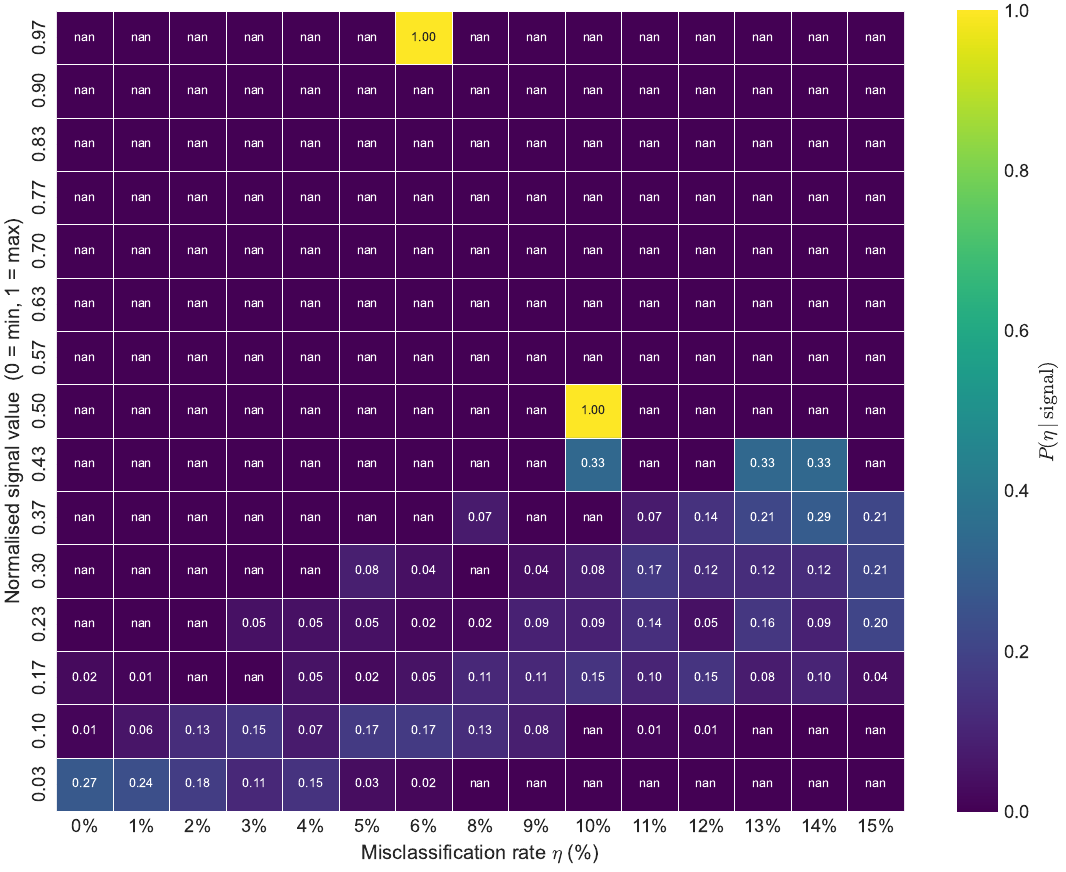}
    \caption{This plot shows the via \ac{lipbvit}SN1 sampled $1 - confidence$ values over the entire experiment as row-normalised 2-D histogram of (signal, $\eta$): given a normalised signal value, each row gives the probability of each misclassification rate. Providing a direct lookup.}
    \label{fig:lookup}
\end{figure}

\newpage
\clearpage

\end{document}